\documentclass[journal]{IEEEtran}
\usepackage{amsmath}
\usepackage{float}
\usepackage{graphicx}
\usepackage{caption}
\usepackage[caption=false]{subfig}
\usepackage{textcomp}
\usepackage{xcolor}
\usepackage{float}
\usepackage{multirow}
\usepackage{csquotes}
\usepackage{url}
\usepackage[colorlinks=true, allcolors=blue]{hyperref}
\usepackage{cite}
\usepackage{tabularx}
\usepackage{booktabs}
\newcolumntype{Y}{>{\raggedleft\let\newline\\\arraybackslash\hspace{0pt}}X}
\newcolumntype{Z}{>{\centering\let\newline\\\arraybackslash\hspace{0pt}}X}

\usepackage{enumitem}
\setlist[itemize]{noitemsep,topsep=0pt}

\ifCLASSINFOpdf

\else

\fi

\begin{document}
\title{Towards an efficient Iris Recognition System on Embedded Devices}
\author{Daniel P. Benalcazar,~\IEEEmembership{Member,~IEEE,}
        Juan E. Tapia,~\IEEEmembership{Member,~IEEE},~Mauricio Vasquez,~Leonardo Causa,~Enrique Lopez Droguett,~Christoph Busch,~\IEEEmembership{Senior Member,~IEEE},
\thanks{Corresponding author: juan.tapia-farias@hda.de}
}

\markboth{Journal of \LaTeX\ Class Files,~Vol.~14, No.~8, October~2022}%
{Shell \MakeLowercase{\textit{et al.}}: Bare Demo of IEEEtran.cls for IEEE Journals}

\maketitle


\begin{abstract}
Iris Recognition (IR) is one of the market's most reliable and accurate biometric systems. Today, it is challenging to build NIR-capturing devices under the premise of hardware price reduction. Commercial NIR sensors are protected from modification. The process of building a new device is not trivial because it is required to start from scratch with the process of capturing images with quality, calibrating operational distances, and building lightweight software such as eyes/iris detectors and segmentation sub-systems. 
In light of such challenges, this work aims to develop and implement iris recognition software in an embedding system and calibrate NIR in a contactless binocular setup. We evaluate and contrast speed versus performance obtained with two embedded computers and infrared cameras. Further, a lightweight segmenter sub-system called "Unet\_xxs" is proposed, which can be used for iris semantic segmentation under restricted memory resources.
\end{abstract}

\renewcommand\IEEEkeywordsname{Keywords}
\begin{IEEEkeywords}
Iris recognition, Iris sensor, NIR Camera, Hardware, embedding systems.
\end{IEEEkeywords}

\section{Introduction}

Iris Recognition (IR) literature has several well-documented and evaluated approaches on how to process iris images, extract the iris code, and perform comparisons \cite{daugman2009iris,bowyer2016handbook,othman2016osiris,he2009efficient,czajka2019domain,gangwar2016deepirisnet,nguyen2017iris,zhao2017towards,minaee2019deepiris,zhao2019deep}. Some of the proposed approaches used classical methods of image processing, such as Gabor Wavelets,  \cite{daugman2009iris, bowyer2016handbook,othman2016osiris}, Localized Binary Patterns (LBP) \cite{he2009efficient}, or Binarized Statistical Image Features (BSIF) \cite{czajka2019domain}, while others used Convolutional Neural Networks (CNN) for segmentation and/or encoding \cite{gangwar2016deepirisnet,nguyen2017iris,zhao2017towards,minaee2019deepiris,zhao2019deep,tapia2021semantic}.
However, only a few previous works aimed to describe how to build a binocular iris imaging device efficiently from scratch at both hardware and software levels. Moreover, a binocular device can also be used as Fitness for Duty system from NIR iris images.

The availability of Raspberry-PI and Jetson-Nano increased the number of devices developed due to low prices and fast adoption compared to regular binocular NIR capture devices.
When an eye detection and segmentation subsystem are deployed on a hardware platform, it requires sufficient amounts of memory to store its parameters (weights and biases), and intermediate computational results exchanged between its deployments. This is not a trivial task. 

In this work, an iris recognition system has been developed from scratch considering the increase of computational power, utilising semantic segmentation, and designing a binocular capture device that makes use of the information of both eyes. The computation power is efficiently allocated by using and comparing two modern embedded boards such as the Raspberry-Pi-4B and the Jetson-Nano. Additionally, we develop a low-weight semantic segmentation network that can work on those boards with fast inference times. Finally, the proposed device arrangement can capture a face image from which both eyes are extracted with good resolution for IR. Therefore, the proposed device could also be used for face recognition as a complementary function.

The contributions of this work are the following:
\begin{itemize}
\item \textit{IR Hardware Design}: the main elements needed to build a NIR iris imaging device with readily available components are described.
\item A comparison between Raspberry-Pi and Jetson-Nano is reported.
\item IR Pipeline: an end-to-end IR pipeline is implemented, which acquires information about the two eyes in a face image. 
\item \textit{Iris Segmentation Network}: a lightweight iris semantic segmentation network is proposed, which finds eyes and segments the iris.
\item \textit{System Calibration and Evaluation}: the optimal distance to the camera is calibrated; additionally, the speed and performance of the proposed system are evaluated. 
\end{itemize}

This paper is organised as follows: related work is described in Section~\ref{sec:related}. The method is outlined in Section~\ref{sec:method}. The databases and data-preprocessing are introduced in section~\ref{sec:datasets}, and the experiments and results are presented in section~\ref{sec:exp_and_res}. Finally,  conclusions are presented in section~\ref{sec:conclusion}.

\section{Related Methods}
\label{sec:related}

\subsection{Iris Recognition Hardware}

There have been a small number of research works that detail how to build IR capture devices from the hardware perspective. Early works focused on the implementation of IR systems on Digital Signal Processors~(DSP) as well as Field Programmable Gate Arrays~(FPGA)\cite{grabowski2006iris,grabowski2011hardware,lopez2011hardware}. Those systems were attractive because they exploited parallel processing to reduce computational time. However, they are expensive in comparison to single-board computers. 

Single-board computers such as Raspberry-Pi offer a small form-factor platform for IR systems. Cruz et al. \cite{cruz2016iris} implemented Daugman's equations on a Raspberry-Pi-2B. 

Bastias et al. \cite{bastias2017method} reported a similar device, but for the purpose of 3D iris scanning. They used a Raspberry-Pi-3B+ with an infrared 8Mpx camera and NIR LEDs. The setup was mounted on a repurposed Virtual Reality (VR) headset. 


Fang et al. \cite{fang2020open} proposed an end-to-end open-source system for IR with PAD. They described how to build an iris capture device using general-purpose electronic components for less than 75 USD. Their system utilises a Raspberry-Pi-3B+, an infrared 5Mpx camera, a NIR optical filter, two NIR LEDs, and control circuits mounted on a breadboard. Fang et al. also produced a complete IR solution with Presentation Atack Detection (PAD) that runs on the restricted computational power of the Raspberry-Pi-3B+.

There have also been some implementations on smartphones. Raja et al. \cite{raja2015smartphone} utilised an iPhone 5S and a Nokia Lumia 1020 as the capture device using Visible spectrum (VS) imaging. The smartphone also produced segmentation inference and recognition \cite{raja2015smartphone}. 

Benalcazar et al. \cite{Benalcazar2019a,Benalcazar2019b} proposed an iris imaging device using VS images, which was used for IR in 3D \cite{benalcazar2020a,benalcazar2020b}. This device was based on a frame similar to Bastias's VR goggles; however, it was built on opaque black acrylic to suppress undesired reflections. The acquisition device was a Samsung S6 Smartphone with a macro lens, capturing close-range images of the iris at 16Mpx resolution. 

While the resolution of the devices of Bastias \cite{bastias2017method} and Benalcazar \cite{Benalcazar2019a, Benalcazar2019b} were greater than that of Fang's device \cite{fang2020open}, the close required proximity of the device to the subject´s eyes made them impractical during the COVID 19 pandemic. Additionally, all of the methods above capture iris images of a single eye, losing the potential performance improvement of using both eyes \cite{sharkas2016neural}.

Efficient iris recognition in Head-mounted embedded displays also has been proposed.
Fadi et al. \cite{fadi-iris1, fadi-iris2, fadi-iris3} developed iris and periocular biometrics for head-mounted displays, including segmentation, recognition, and synthetic data generation focusing on Augmented Reality (AR) and Virtual Reality (VR) applications.

\subsection{Iris Semantic Segmentation}

Recent works in IR use deep learning to segment and localise the pupil, and the iris in a periocular image \cite{fadi-iris4, ronneberger2015unet, mishra2019ccnet, tapia2021semantic}. 
The Criss-Cross Attention Network (CCNet), developed by Mishra et al. \cite{mishra2019ccnet}, is an iris semantic segmentation network based on U-Net \cite{ronneberger2015unet}. This lightweight network was trained by Fang et al. \cite{fang2020open} to predict a binary mask of the iris from NIR images. They utilised the Hough transform to localise the pupil and the iris from the binary mask. CCNet was explored and re-trained by Tapia et al. \cite{tapia2021semantic} to work with highly dilated images under the influence of alcohol. They also developed two faster localisation algorithms that had better performance than the Hough transform: Least Means Squares (LMS) and Mixed \cite{tapia2021semantic}.

Tapia et. al also developed DenseNet10 \cite{tapia2021semantic}, a reduced version of DenseNet101 \cite{iandola2014densenet} that can semantically segment the pupil, the iris, the sclera, and the background \footnote{\url{https://github.com/Choapinus/DenseNet10}}. This network obtained greater performance than CCNet on alcohol images at the expense of doubling the number of parameters. The centre of mass was used to localise the ellipses of the pupil and the iris.

CCNet and DenseNet10 are the main networks used in this work. However, there is abundant literature on semantic segmentation networks. For instance, Wu and Zhao \cite{wu2019study} developed a segmentation network based on U-Net with good performance. Sip-SegNet \cite{hassan2020sip} employs Convolutional Neural Networks along with adaptive fuzzy filtering techniques to segment the pupil iris and the sclera. Wang et al. \cite{wang2022light} proposed a lightweight network for iris semantic segmentation. Osorio et al. \cite{osorio2017semantic, osorio2018visible} developed a multi-class segmentation network for VS images. Li et al. \cite{li2021robust} developed a segmentation network that works in uncooperative scenarios.

\section{Methods}
\label{sec:method}

\subsection{Device Design}

\begin{figure}[tb]
    \centering
    \captionsetup{justification=centering}
    \includegraphics[width=1\linewidth]{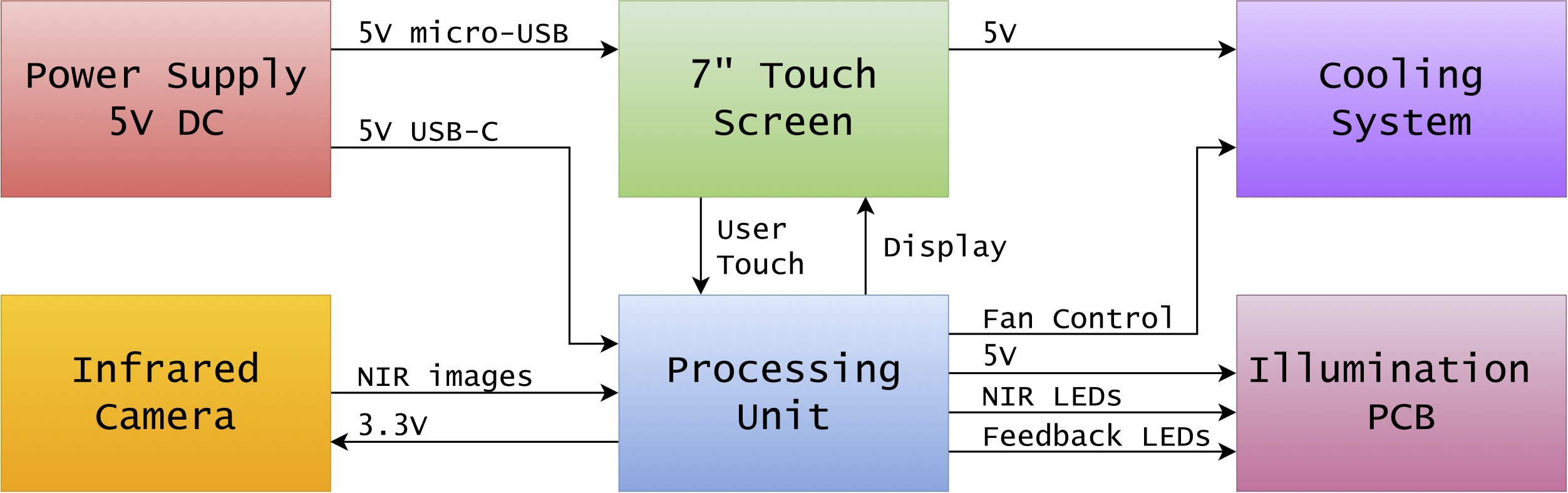}
    \caption{Block diagram of the Hardware modules.}
    \label{fig:block}
\end{figure}

The proposed iris-capturing device is designed to be portable, light and contactless, acquire NIR images of both irises, show relevant information to the user, and perform IR on its own. According to previous statements, the device is composed of the following modules: Processing unit, Cooling system, Capture Device, Illumination Printed Circuit Board (PCB), Display module, Power supply, and Case. 
All the modules are interconnected, as illustrated in the block diagram of Figure~\ref{fig:block}. 

The processing unit is in charge of commanding the rest of the modules and running the IR software. We compare the Raspberry-Pi-4B and Nvidia's Jetson-Nano for the processing unit. Both boards are readily available, have a small form factor and permit controlling peripheral devices. 

The Raspberry-Pi-4B is based on a quad-core ARM-A72 64-bit CPU that runs at 1.5GHz. The model used in this work has 8GB of RAM and no CUDA-compatible GPU. 

On the other hand, The Jetson-Nano is based on a quad-core ARM-A57 64-bit CPU that runs at 1.43GHz. It has 4GB of RAM and a 128-core Maxwell CUDA-compatible GPU. Therefore, the Raspberry-Pi has a better CPU and RAM, but the Jetson-Nano has a GPU that can run neural networks. We will analyse the trade-offs of each in terms of computational performance in Section~\ref{sec:exp_and_res}.

The cooling system is different for the two choices of processing unit. For the Raspberry-Pi-4, passive heatsinks were installed on the main chips, and a fan was installed in the case. The fan blows cool air to the heatsinks, and hot air exits through holes at the sides of the case. The fan is turned on and off automatically from the OS through the General Purpose Input Output (GPIO) pins and a current amplification transistor. On the other hand, a single heatsink, installed by default, cools the main components of the Jetson-Nano. A 4-pin fan was installed to cool the heatsink, and it is controlled by the OS with Pulse Width Modulation (PWM). Both options effectively remove the excess heat from the device and prolong the lifespan of the device. 

The display module is a seven-inch touchscreen for the operator's interaction with the capture system. While capturing the eyes, it shows the biometric attendant a mirror of their actions so they can centre the eyes of the captured subject inside the frame. The display also tells the subjects when the capture process has started and ended. Finally, it shows a welcome screen if the biometric attendant was identified or instructions to follow if otherwise. Touch information can be used to start the capture process or wake up the device if inactive for a long time.

\begin{figure}[t]
    \centering
    \captionsetup{justification=centering}
    \includegraphics[width=0.75\linewidth]{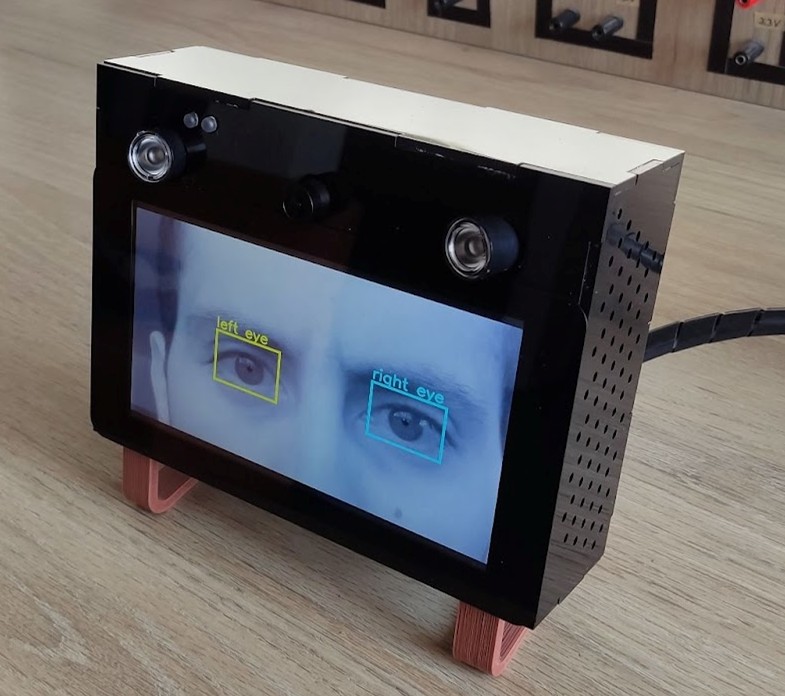}
    \caption{Image of the assembled device.}
    \label{fig:device}

    \vspace{5mm}

    \begin{center}
        \subfloat[Cam v1 5Mpx \label{subfig:cam_v1}]{{\includegraphics[width=0.48\linewidth]{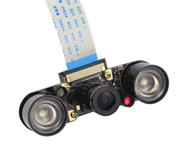} }}
        \subfloat[Cam v2 8Mpx \label{subfig:cam_v2}]{{\includegraphics[width=0.48\linewidth]{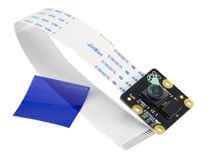} }}
    \end{center}
    \caption{NoIR camera modules for the Raspberry-Pi.}%
    \label{fig:cams}%
\end{figure}

Finally, the case holds all components together and facilitates cooling. A simple squared design was fabricated on laser-cut black acrylic. All the modules are secured to the acrylic frame by means of 3D-printed parts, nuts and screws. The bottom part of the device has a universal nut that can be mounted on tripods of any size. 
An image of the assembled device can be seen in Figure~\ref{fig:device}. It also illustrates that the camera input is a Face Region of Interest (ROI) with two eyes.

The setup of the capture device and illumination PCB was based on \cite{fang2020open}. For the capture device, we compare the 5Mpx NoIR camera module v1 and the 8Mpx NoIR camera module v2 for Raspberry-Pi, as shown in Figure~\ref{fig:cams}. The illumination PCB has two NIR LEDs to illuminate the eyes and two VL LEDs that give feedback to the user. The four LEDs are controlled by the GPIO pins of the processing unit by means of transistor-based current amplification. This enhances the brightness of the LEDs while avoiding drawing to much current from the GPIO. 
The main differences with the literature are the following: (i) It was designed and fabricated a custom PCB for the transistors that control the LEDs instead of using a breadboard; (ii) since PAD is not an objective for this work, both NIR LEDs are turned on and off at the same time instead of individually; (iii) we did not use a NIR optical filter since preliminary tests showed that the filter's image quality drops at night; and (iv) It was captured a face image and extract both eyes instead of capturing a single eye in close proximity.

\subsection{Semantic Segmentation Network}

It was developed as a reduced version of U-Net \cite{ronneberger2015unet}, which is called UNet\_xxs, for the purpose of iris semantic segmentation under limited computational power. Iteratively, we removed layers from CCNet's version of U-Net \cite{mishra2019ccnet, fang2020open}, trained and tested performance until a small network with acceptable performance was obtained. The result of this pruning process is an architecture with only 28,146 parameters. Considering that CCNet \cite{mishra2019ccnet, fang2020open} has 122,514 and DenseNet10 has 210,732 parameters, the proposed UNet\_xxs network represents a parameter reduction of 77\% and 93\% respectively. 

\begin{figure}[b]
    \begin{center}
        \subfloat[Find eyes \label{subfig:find_eyes}]{{\includegraphics[width=0.50\linewidth]{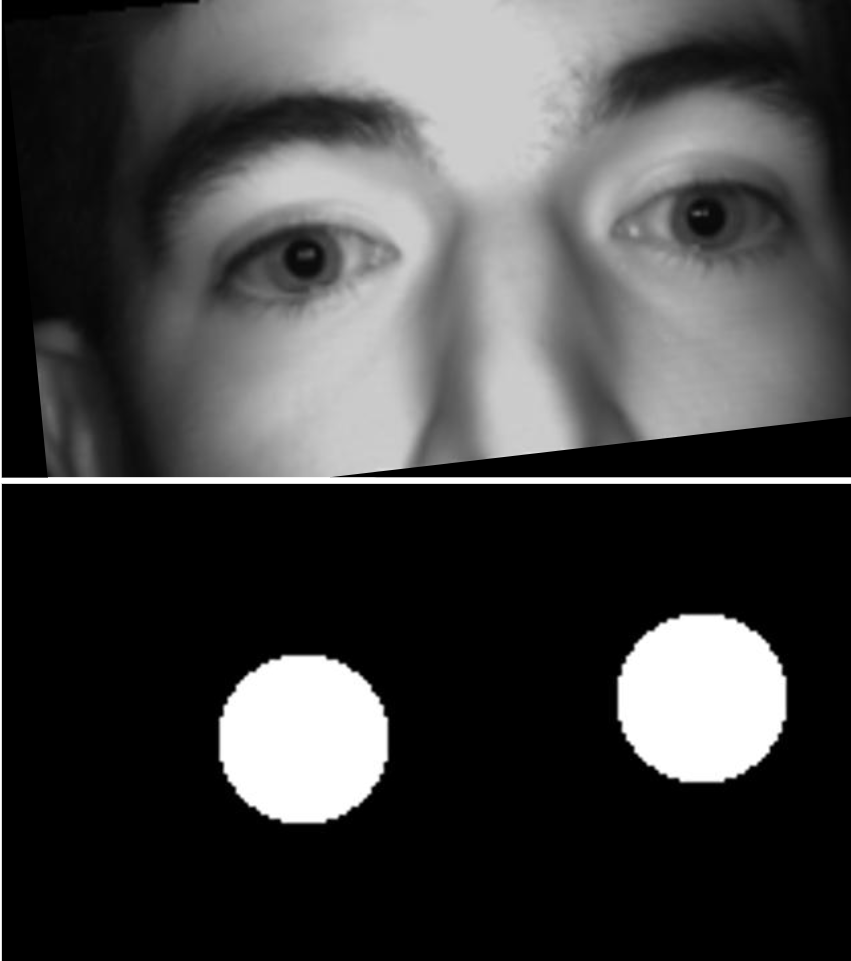} }}
        \subfloat[Segment iris \label{subfig:segment_iris}]{{\includegraphics[width=0.40\linewidth]{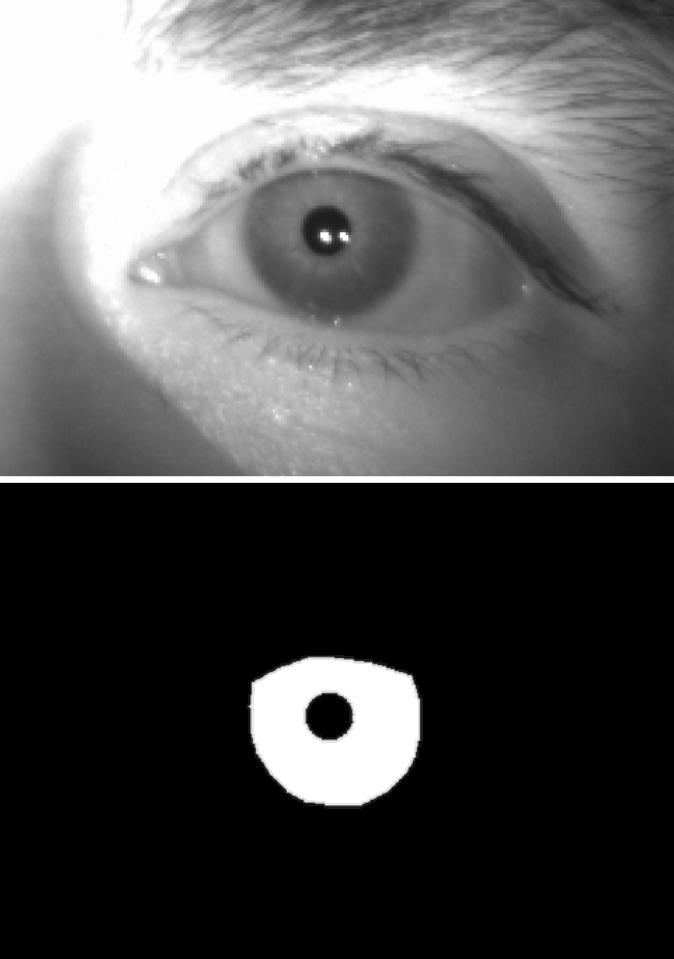} }}
    \end{center}
    \caption{Illustration of the two trained tasks. Top: input image. Bottom: segmentation mask.}%
    \label{fig:tasks}%
\end{figure}

\begin{figure*}[t]
    \centering
    \includegraphics[width=\linewidth]{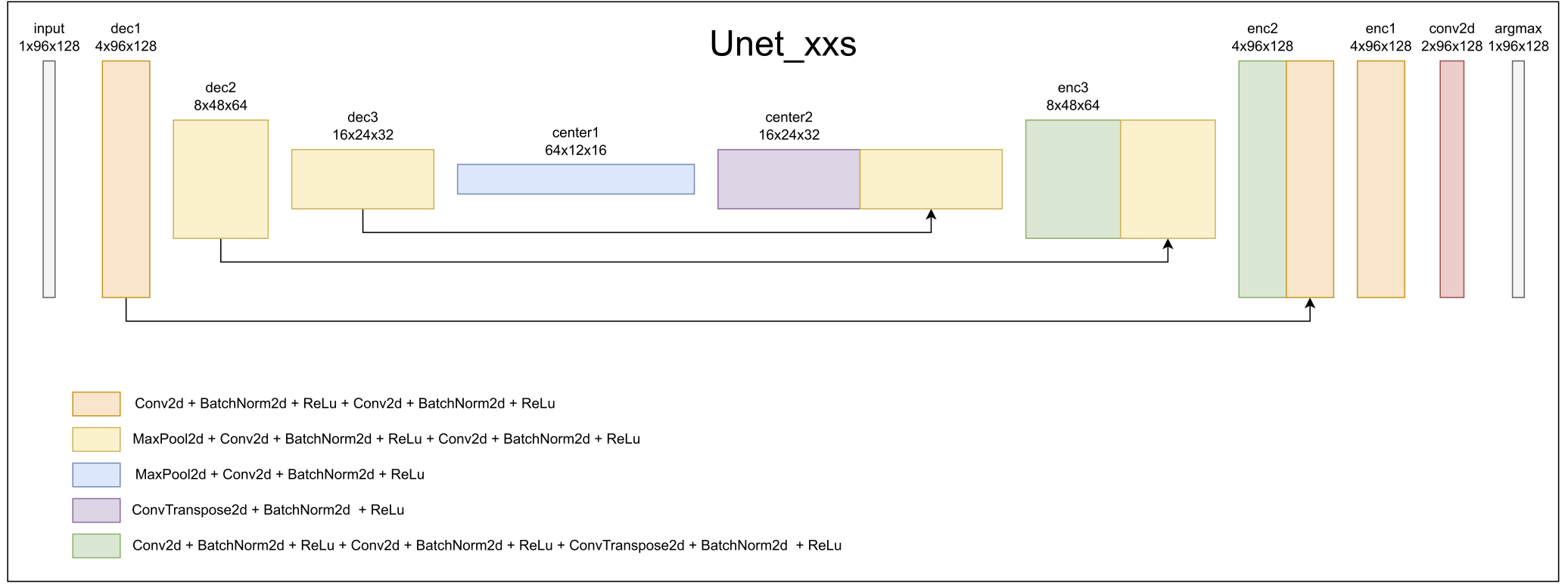}
    \caption{Architecture of UNet\_xxs network.}
    \label{fig:Unet_xxs}
\end{figure*}

We trained CCNet and UNet\_xxs on two tasks using the dataset from \cite{tapia2021semantic}. In the first task, the networks have to find the two eyes from the face-ROI image, as illustrated in Figure~\ref{subfig:find_eyes}. This is used to crop the two periocular images of the left and right eyes. The second task is to find the iris region in a periocular image, as depicted in Figure~\ref{subfig:segment_iris}. The two tasks were trained separately and handled by two independent models. The architecture of UNet\_xxs is illustrated in Figure~\ref{fig:Unet_xxs}.

\begin{figure*}[t]
    \centering
    \includegraphics[width=\linewidth]{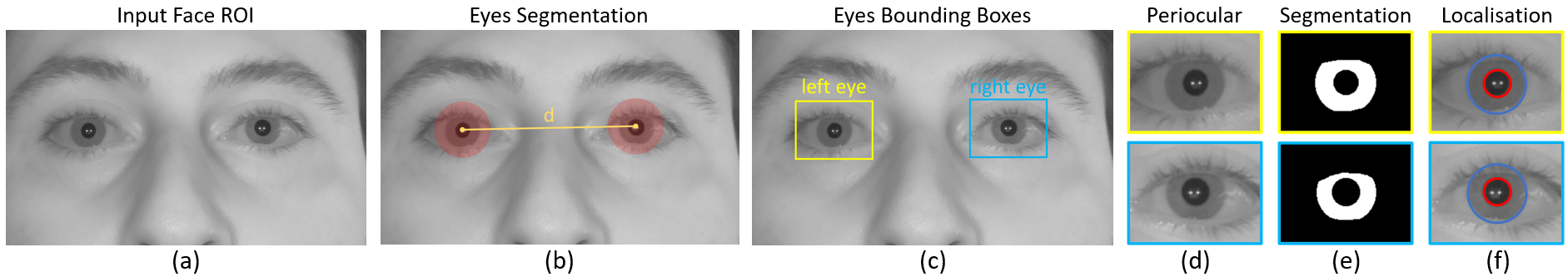}
    \caption{Iris segmentation and localisation pipeline.}
    \label{fig:pipeline}
\end{figure*}


The segmentation and localisation pipeline is shown in Figure~\ref{fig:pipeline}. It illustrates how the input Face ROI image (Figure~\ref{fig:pipeline}\textcolor{blue}{a}) is processed to find, crop, segment and localise the left and right irises. The output segmentation masks, as well as the localisation coordinates, are essential for IR.

First, a segmentation network (CCNet or UNet\_xxs) is used to find the eyes. The network takes the Face ROI image (Figure~\ref{fig:pipeline}\textcolor{blue}{a}) and produces a segmentation mask where two big circles indicate the position of the eyes, as highlighted in red in Figure~\ref{fig:pipeline}\textcolor{blue}{b}. Then the distance $d$ between the centroids of both circles are computed. Next, the bounding boxes of each eye are found (Figure~\ref{fig:pipeline}\textcolor{blue}{c}). These are rectangles placed at the centre of each circle with width $0.44 \times d$ and height $0.33 \times d$. This box size was determined empirically based on component and display sizes. After that; the bounding boxes are used to crop each eye and obtain the periocular images (Figure~\ref{fig:pipeline}\textcolor{blue}{d}).

Then, the segmentation networks take the periocular images as inputs and return the segmentation masks (Figure~\ref{fig:pipeline}\textcolor{blue}{e}), where the iris region is highlighted in white. Finally, the circles of the two pupils and irises are found (Figure~\ref{fig:pipeline}\textcolor{blue}{f}) using the LMS and the mixed algorithms proposed by Tapia et al. \cite{tapia2021semantic}. The results are the $(x,y,r)$ triplets of each circle, where $(x,y)$ are the coordinates of the circle's centre and $r$ is the radius.

\subsection{Iris Recognition}

For IR, we used the method of Czajka et al. \cite{czajka2019domain}, implemented on the Raspberry-Pi by Fang et al. \cite{fang2020open}. This method takes a periocular iris image, and the localisation $(x,y,r)$ coordinates to produce a polar-coordinate normalised iris image, also known as rubber sheet \cite{daugman2009iris}. The rubber sheet has a size of $1 \times 64 \times 512$. A rubber sheet of the segmentation mask is also generated. Then, the BSIF-based ICA filters \cite{czajka2019domain} is applied to encode the iris rubber sheet. The ICA filters are 7 filters of $15 \times 15$ and are available on open source by \cite{fang2020open}. Finally, two encoded irises are compared using the Modified Hamming Distance, as described in Section~\ref{sec:metrics}. The masks omit information from non-usable areas such as the pupils, eye leads and eyelashes from the comparison score computation.

\section{Datasets}
\label{sec:datasets}

\subsection{Segmentation Datasets}

Both segmentation networks (CCNet and UNet\_xxs) were trained for two separate tasks: finding eyes and segmenting the iris. To solve the first task using semantic segmentation, we used the face ROI images of \cite{tapia2021semantic}, face images from the FRCG dataset, as well as our own captured images using the Raspberry-Pi. The ground-truth (GT) masks of each face and face ROI image were created by placing two big circles on each eye, as seen in Figure~\ref{subfig:find_eyes}. For this purpose, the pupil centres were marked on each image, and the pupil-to-pupil distance $d$ was measured. Then, two circles with radius $0.2\times d$ were placed at the pupil centres of each image. Additionally, we trained the network to produce the circles even though the eyes were closed, so we placed images of closed eyes in the training set. Finally, the network was trained to leave the image in black if no eyes were present on the input image. To accomplish this, we added natural images as backgrounds in the training set. 

\begin{figure}[t]
    \centering
    \includegraphics[width=\linewidth]{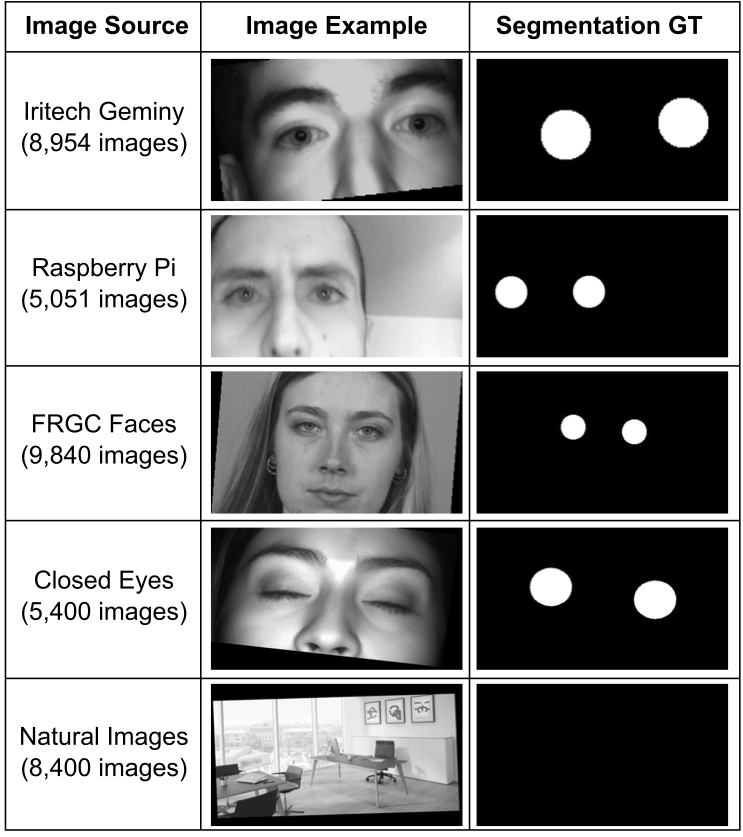}
    \caption{Find Eyes Dataset.}
    \label{fig:find_eyes_data}
\end{figure}

Figure~\ref{fig:find_eyes_data} shows the number of images used for each source, a sample image and its GT segmentation mask. The number of images in Figure~\ref{fig:find_eyes_data} accounts for offline data augmentations of rotation that were performed on each image. The total number of images was 37,645, and it was divided into 30,116 for training, 3,764 for validation and 3,765 for testing. 

On the second task, the networks were trained to segment the iris region in a periocular image. Tapia et al.'s dataset \cite{tapia2021semantic} had GT of semantic segmentation for the pupil, the iris and the sclera. In this work, we isolated the iris label to train the networks only on the iris region. Additionally, we segmented the captured Raspberry-Pi images using the pre-trained DensNet10 \cite{tapia2021semantic} and used the outputs of that network as the ground truth masks. 

Similarly to the first task, we trained the iris segmentation network to output a black image when no iris was present in the input image. For this purpose, we placed natural images, closed eyes and face close-ups in the dataset. 

\begin{figure}[t]
    \centering
    \includegraphics[width=\linewidth]{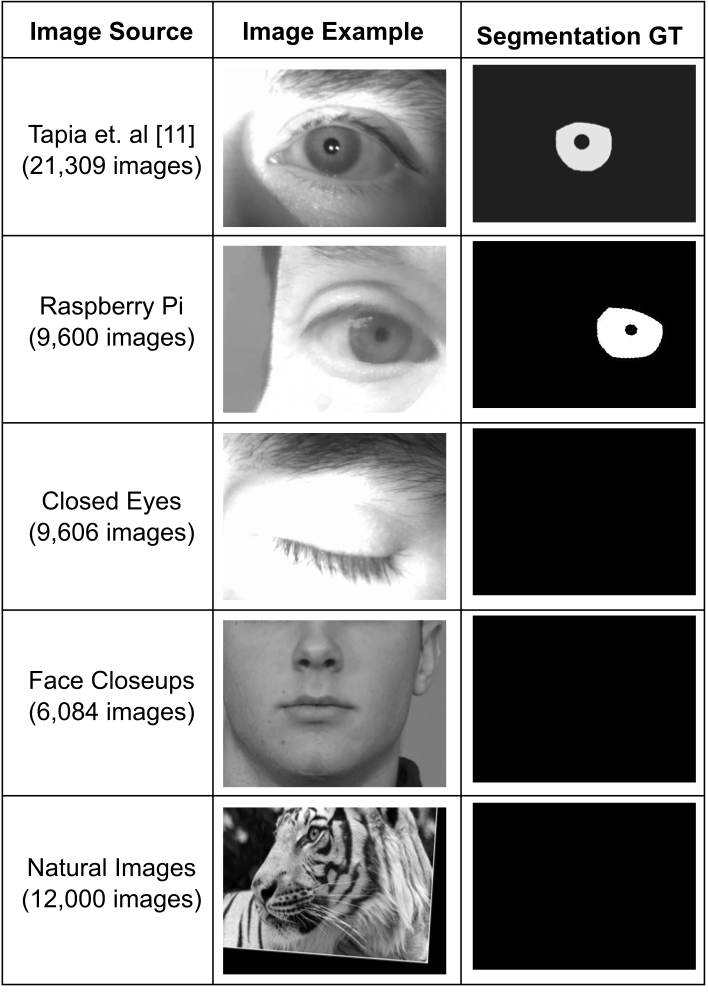}
    \caption{Segmentation Dataset.}
    \label{fig:seg_data}
\end{figure}

Figure~\ref{fig:seg_data} illustrates the number of images, an example and the corresponding segmentation mask per source. The number of images considers offline data augmentation of random rotations placed on all the sources. The total number of images is 58,599 and was divided into 44,748 images for training, 7,990 for validation and 5,861 for testing.

It is worth pointing out the fact that the first network can find and crop closed eyes while the second is not able to do so. This in turn helps count the number of times a person blinks as well as measure the blinking frequency when the networks perform continuous predictions on image sequences.

\subsection{Iris Recognition Datasets}

The dataset of Tapia et al.'s \cite{tapia2021semantic} has a great number of face ROI images and iris images with segmentation ground truth, so it is ideal for training segmentation networks. However, it is not a good dataset for iris recognition since it has image sequences of around 75 frames per subject, and subjects might appear on alcohol and non-alcohol sets with different IDs since images were captured with different sensors. Therefore, we chose to use different datasets for IR tests.

For fast and iterative IR tests, we used the dataset of Benalcazar et al. \cite{Benalcazar2019a}. It has 96 subjects, and 5 NIR images of the right eye of each subject were captured, for a total of 480 images. Therefore, there are 960 mated and 114,000 no-mated comparisons in the dataset. The average image width is 388.4px, and the aspect ratio is 4:3.

For a final IR benchmark evaluation, we used the Notre Dame LG4000 dataset \cite{UND-data}. 
From the original dataset, we removed low-quality images and images with a predominant nose in the frame. The remainder of the 10,959 images were classified between the Left and Right eyes. We have different IDs on the left and right images of the same individual since the irises are different. In total, there were 811 IDs, which were partitioned in a person-disjoint manner, as 487 for training, 162 for testing, and 162 for validation. The partition corresponds to 6,663 train images, 2,090 test images and 2,203 validation images. The test set has 19,029 mated comparisons and 105,470 non-mated comparisons. We named this partitioned dataset as ND-LG400-LR and made the partition lists available on GitHub\footnote {\url{https://github.com/TOC-SR-226/ND-LG4000-LR}}. Additionally, we generated the segmentation GT by segmenting the 10,959 images with DenseNet10 and correcting mistakes by hand.

\section{Metrics}

\subsection{Modified Hamming Distance}
\label{sec:metrics}

For comparing two binary iris codes, one of the most used metrics is the Modified Hamming Distance (HD). The two encoded irises ($codeA$ and $codeB$) and the rubber sheets of the segmentation masks ($maskA$ and $maskB$) are used to compute the HD \cite{daugman2009iris}, as described in \eqref{eq:HD}. The smaller the $HD$ value, the more similar the two irises are.

\begin{equation}
    HD = \frac{||(codeA \otimes codeB) \cap maskA \cap maskB||}{||maskA \cap maskB||}
\label{eq:HD}
\end{equation}

\subsection{D-prime Score}

The $d'$ score evaluates how separated the mated and non-mated comparison distributions are. It is computed score using \eqref{eq:d_prime} \cite{daugman2009iris}, in which $\mu_1$ and $\mu_2$ are the mean of the mated and non-mated score distributions respectively, and $\sigma_1$ and $\sigma_2$ are the corresponding standard deviations. The greater the ($d'$) value, the greater the separation between the two distributions.

\begin{equation}
    d' = \frac{|\mu_1 - \mu_2|}{\sqrt{0.5 \times (\sigma_1^2 + \sigma_2^2)}}
\label{eq:d_prime}
\end{equation}

\subsection{Equal Error Rate}

The Equal Error Rate (EER) in a biometric test is the operating point at which the False Match Rate (FMR) and the False Non-Match Rate (FNMR) are equal. The smaller the EER value is, the less overall classification error the system has. 

\subsection{Signal to Noise Ratio}

The Signal to Noise Ratio (SNR) is the ratio between a signal's amplitude and the noise's standard deviation. In this work, we are interested in evaluating the SNR of iris images under different conditions. Therefore, we compute the SNR using \eqref{eq:SNR}, where the amplitude of the signal is the difference between the mean intensity values of the iris and the sclera, and the noise is considered as the standard deviation of the sclera. That is because the sclera is normally white; thus, any random variation of intensity from one pixel to the next is due to camera noise artefacts.

\begin{equation}
    SNR = \frac{A}{\sigma} = \frac{|\mu_{iris} - \mu_{sclera}|}{\sigma_{sclera}}
\label{eq:SNR}
\end{equation}

\subsection{Intersection Over Union}
The bitwise Intersection over Union (IoU) \cite{tapia2021semantic} metric is used to evaluate semantic segmentation accuracy. It compares the ground truth and predicted segmentation masks ($M_A$ and $M_B$) by means of \eqref{eq:IoU}. This equation counts the number of bits that are high in a logical AND operation and divides it by the number of logical ones in the logical OR operation between the two masks. 

\begin{equation}
IoU=\frac{\sum_{}^{} (M_A \wedge M_B)}{\sum_{}^{}  (M_A \vee M_B) +  \epsilon}
\label{eq:IoU}
\end{equation}

The range of values of the IoU is between 0 and 1, where 1 is a perfect match. A small positive value $\epsilon$ is added to the denominator to avoid division by zero.

\section{Experiments and Results}
\label{sec:exp_and_res}

Experiments 1, 2, 3 and 4 aim to resolve a crucial aspect in the fabrication of an iris imaging device, which is finding the optimal distance between the camera and the subject in terms of image resolution, SNR and available iris area. Since the v1 and v2 camera modules used in this work cannot regulate the focal distance automatically, the focus has to be manually adjusted upon assembly. Therefore, the most crucial factor to find is the optimal distance between the subject and the camera, at which to fix the focal plane. 


Experiments 5, 6 and 7 evaluate the performance of the proposed segmentation network and pipeline in terms of accuracy and speed.

Finally, Experiment 8 evaluates the overall IR performance on the ND-LG400-LR dataset.

\subsection{Experiment 1: IR Performance vs Image Resolution}

First, we explore the minimum resolution at which IR still has a reasonable performance. For this purpose, we use the dataset of \cite{Benalcazar2019a}, which is composed of 480 NIR images of the right eyes of 96 subjects. In this test, we varied the image width of each image from 388.4 px down to 25 px in 11 steps and measured the mean iris radius on each step. For each resolution, we also performed IR tests using the method of \cite{czajka2019domain}, and scored the EER, as well as the $d'$ value. 

\begin{figure}[t]
    \centering
    \captionsetup{justification=centering}
    \includegraphics[width=0.8\linewidth]{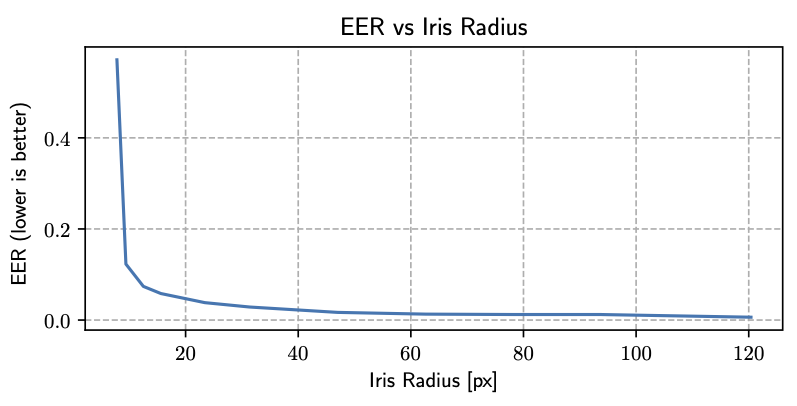}
    \caption{EER in IR decreases when iris radius increases.}
    \label{fig:EER_vs_r}
    
    \vspace{3mm}
    
    \centering
    \captionsetup{justification=centering}
    \includegraphics[width=0.8\linewidth]{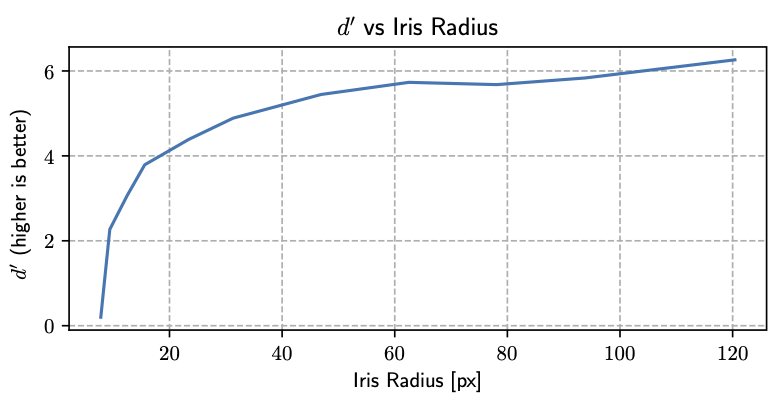}
    \caption{In IR tests, $d'$ increases with iris radius.}
    \label{fig:dp_vs_r}
\end{figure}

Figure~\ref{fig:EER_vs_r} and Figure~\ref{fig:dp_vs_r} show the relationship between the iris radius and the IR metrics. The elbow in Figure~\ref{fig:EER_vs_r} occurs at an iris radius of 9.2px; however, EER stabilises at 46.9px. On Figure~\ref{fig:dp_vs_r}, the $d'$ also presents less rate of growth over 45px of iris radius. 

\begin{figure}[b]    
    \centering
    \subfloat[ \label{subfig:dist_max}]{{\includegraphics[width=0.32\linewidth]{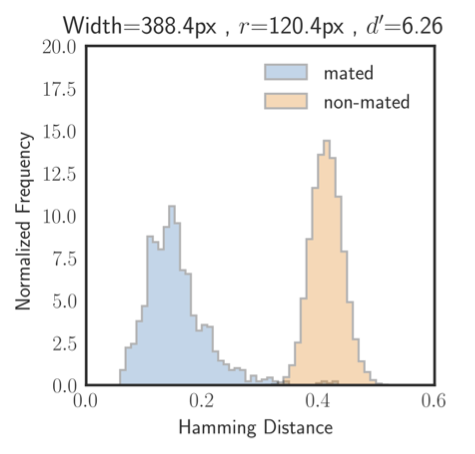} }}%
    \subfloat[ \label{subfig:dist_min}]{{\includegraphics[width=0.32\linewidth]{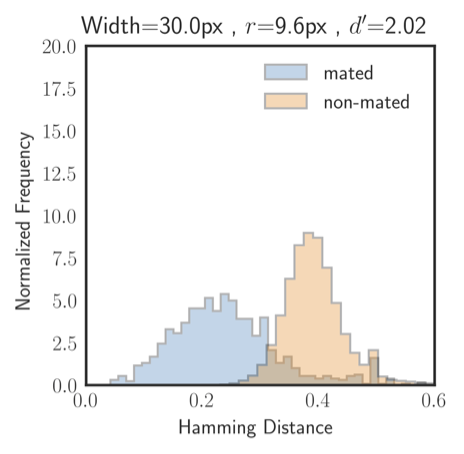} }}
    \subfloat[ \label{subfig:det}]{{\includegraphics[width=0.32\linewidth]{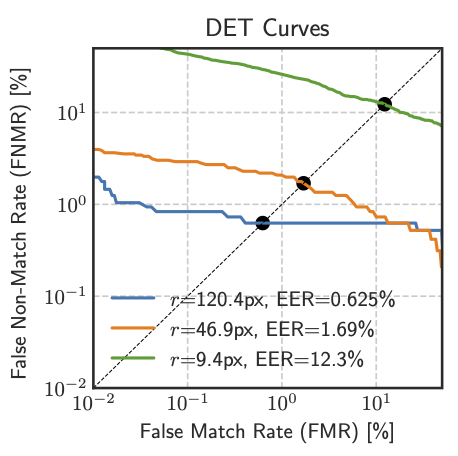} }}
    \caption{IR tests at different resolutions.}%
    \label{fig:distributions}%
\end{figure}

Figure~\ref{fig:distributions} shows in more detail the IR tests at three different resolutions when the iris radius is 120.4 px, 46.9 px, and 9.8 px. It can be seen that there is much more overlap at 9.2 px (Figure~\ref{subfig:dist_min}) compared to the maximum resolution 120.4px (Figure~\ref{subfig:dist_max}); that is why $d'$ diminishes from 6.26 to 2.02. The Detection Error Trade-off (DET) curves in Figure~\ref{subfig:det} reveal the system behaviour for the three resolutions. It shows the expected result, that errors decrement with higher resolutions; however, at an iris radius of 46.9 px, the EER increases only by 1\% with respect to the maximum resolution.

Therefore, we take the value of 45px as the minimum iris radius for an acceptable IR recognition performance.

\subsection{Experiment 2: Iris Radius vs Distance to Camera}

Now that a minimum iris radius of 45px has been established, we must correlate the iris radius with the distance to the camera. For this test, on the device, we captured 4 face images at distances of 10, 15, 20, 25, 30, 40, 50, 60 and 70 cm from the camera. Then, the iris radius on the 8 eyes contained in the 4 frames are manually measured, and the mean iris radius is scored for each distance value. We repeated this test three times with the following combination of cameras and processors: Raspberry-Pi with 5Mpx camera v1, Raspberry-Pi with 8Mpx camera v2, and Jetson-Nano with 8Mpx camera v2. The OpenCV \cite{opencv_library} libraries used in the Raspberry-Pi program can only acquire images of 1,920 $\times$ 1,080 px from the cameras; however, the Jetson-Nano was able to read images of 3,264 $\times$ 1,848 px using Gstreamer. 

\begin{figure}[t]
    \centering
    \captionsetup{justification=centering}
    \includegraphics[width=1.0\linewidth]{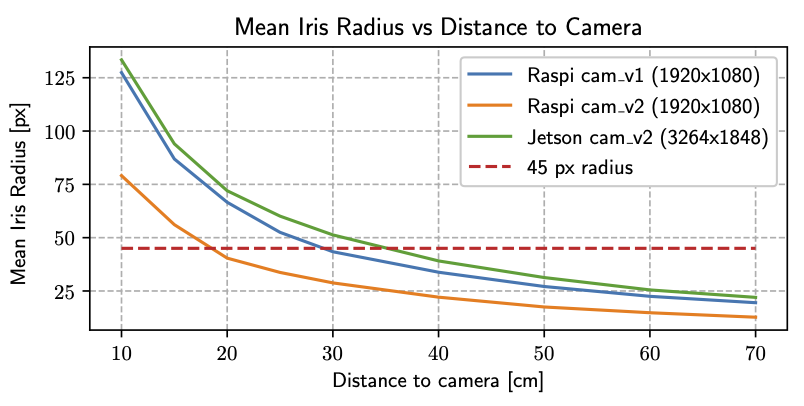}
    \caption{In IR tests, d' increases with iris radius.}
    \label{fig:r_vs_d}
\end{figure}

Figure~\ref{fig:r_vs_d} shows the relationship between the iris radius and the distance to the camera for each device. A red dashed line marks the 45px minimum iris radius found in the previous test. The Raspberry-Pi with the camera v1 has a similar curve to that of the Jetson-Nano with the camera v2 module. The intersection with the 45 px line occurs at 30cm and 35cm from the camera, respectively. Therefore, to work with an iris radius over 45 px, the distance to the camera must not exceed 35 cm.

According to Figure~\ref{fig:r_vs_d}, the Raspberry-Pi gets a lower resolution with the camera v2 module than using the camera v1 module at the same distance to the camera. That is because the camera v1 module possesses a magnification lens, as shown in Figure~\ref{subfig:cam_v1}. However, the Jetson-Nano is capable of getting even better iris resolutions than the Raspberry-Pi with the camera v1 module for the same distances as the camera without the magnification lens. The Jetson-Nano does it at the expense of working with higher-resolution images. Thus, we decided to use the camera v1 module exclusively with the Raspberry-Pi, and the camera v2 module with the Jetson-Nano in further tests.

\vspace{-0.3cm}

\subsection{Experiment 3: Signal to Noise Ratio}

In this experiment, we analyse the SNR at increasing distances from the camera. For this test, we used the camera v2 module with the Jetson-Nano. One image of the subject was taken at 10, 15, 20, 25, 30, 40, 50, 60 and 70cm from the camera, and the iris and the sclera of both eyes were manually segmented on each image. SNR was computed using \eqref{eq:SNR} for each distance value.

\begin{figure}[htb]
    \centering
    \captionsetup{justification=centering}
    \includegraphics[width=0.8\linewidth]{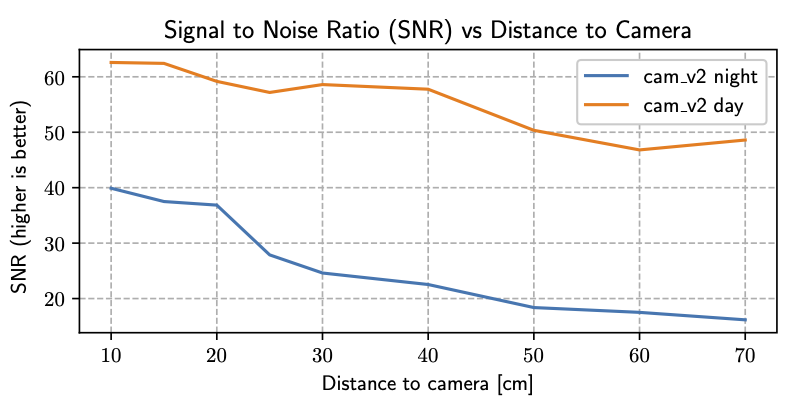}
    \caption{Signal to Noise Ratio at several camera distances.}
    \label{fig:snr}
    
    \vspace{3mm}
    
    \centering
    \captionsetup{justification=centering}
    \includegraphics[width=0.8\linewidth]{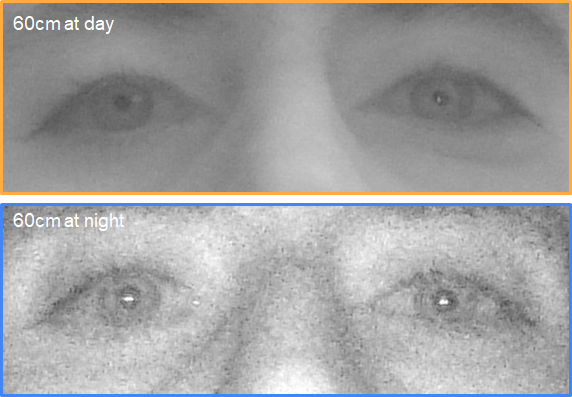}
    \caption{Two Face ROI images at 60cm from the camera. Top: captured during the day. Bottom: captured at night.}
    \label{fig:snr_eyes}
\end{figure}

In total, two capture sessions were taken, one during the day and one at night. This test can be seen in Figure~\ref{fig:snr}. 

In general, SNR diminishes when the distance to the camera increases both at day and night. However, SNR values are much greater during the day than at night. This means that at night the incident NIR light is not enough and the camera has to compensate for the gain, exposure and contrast at the expense of adding extra noise. Figure~\ref{fig:snr_eyes} illustrates how at night, there is much more camera noise present in the image. According to the day curve in Figure~\ref{fig:snr}, over 40 cm of distance, the SNR starts to drop more rapidly. Therefore, the maximum distance with good SNR is 40 cm, according to this test. 

\subsection{Experiment 4: Available Iris Area regarding Subject's Gaze}

\begin{figure}[htb]
    \centering
    \captionsetup{justification=centering}
    \includegraphics[width=0.9\linewidth]{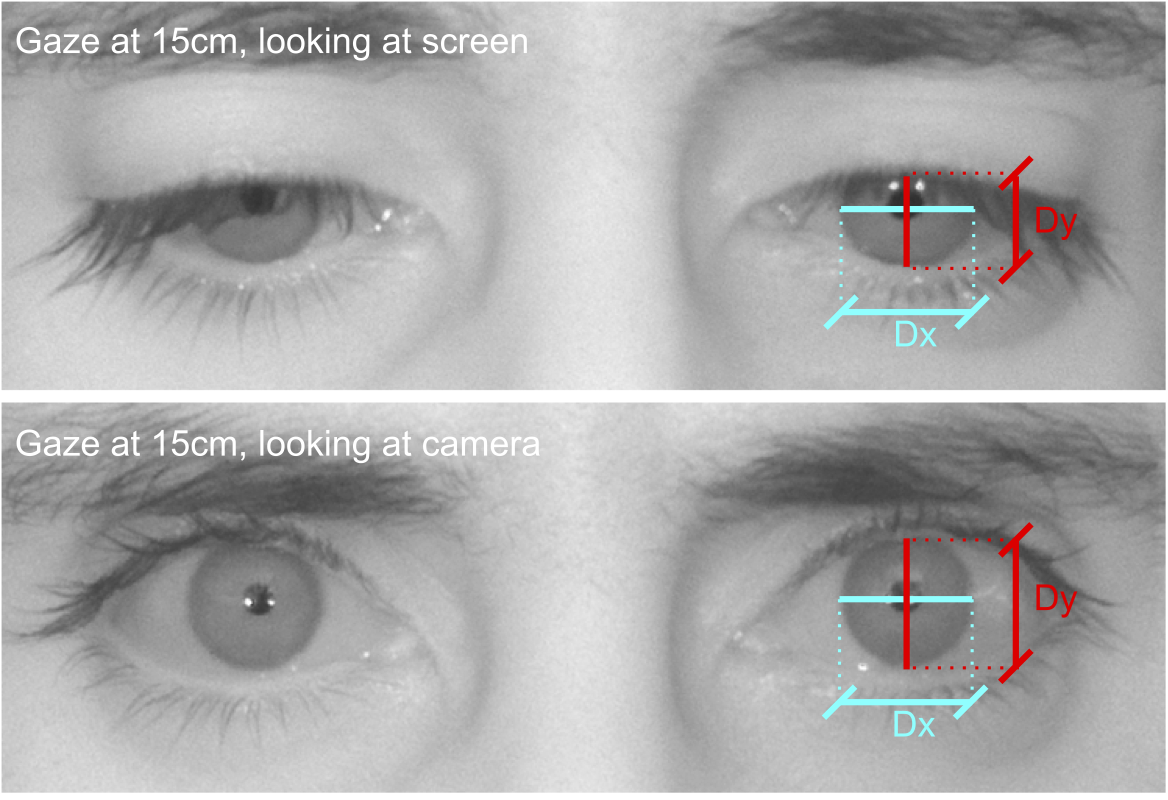}
    \caption{Gaze analysis at 15cm from the camera. Top: looking at the screen. Bottom: looking at the camera.}
    \label{fig:gaze_eyes}
    
    \vspace{3mm}
    
    \centering
    \captionsetup{justification=centering}
    \includegraphics[width=1.0\linewidth]{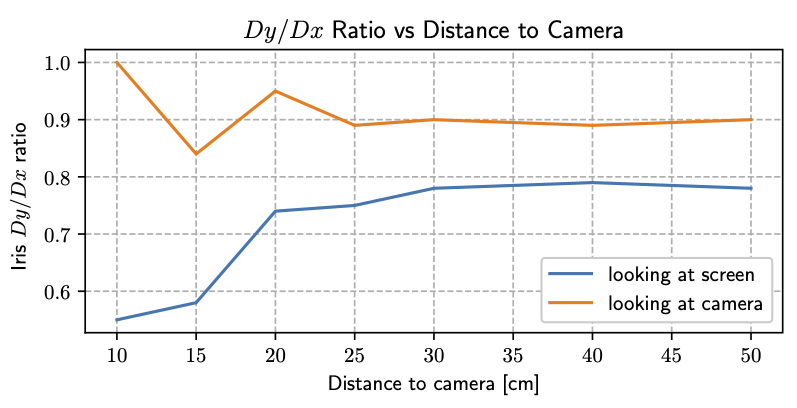}
    \caption{Eye aperture when looking at the camera or the screen for different distances to the camera.}
    \label{fig:gaze}
\end{figure}

An important aspect to take into account is that the users might not look straight into the camera, but instead be distracted by the screen. Despite the camera and the screen being physically close, the subjects' gaze is different when looking at the camera than at the screen, as illustrated in Figure~\ref{fig:gaze_eyes}. When looking at the camera, more iris surface is present in the image, whereas when the subject directs their gaze downwards, eye leads and eyelashes obstruct the iris. We conducted the following experiment to quantify the amount of available iris surface. For the distances to the camera of 10, 15, 20, 25, 30, 40 and 50cm, an image was captured when the subject looked directly at the camera and another when looking at the screen. Then, the iris diameter along the $x$ and $y$ axis was measured for each eye (see Figure~\ref{fig:gaze_eyes}), and the mean ratio of $Dy/Dx$ between the left and the right eye was scored for each frame. This ratio represents how much of the iris was occluded due to eye leads and eyelashes in the $y$ axis with respect to the $x$ axis, where the iris is never occluded. The results of this test are shown in Figure~\ref{fig:gaze}. When looking at the camera, the ratio settles at around 0.9 for distances over 25 cm. However, when looking at the screen, the ratio stabilises slightly below 0.8 from a distance of 30 cm. At a distance between 20 and 25 cm, the ratio is around 0.75, which might also be acceptable. Therefore, according to this test, the optimal distance to the camera is above 25 cm.

\subsection{Analysis of Experiments 1 through 4}

\begin{figure}[b]
    \centering
    \captionsetup{justification=centering}
    \includegraphics[width=1.0\linewidth]{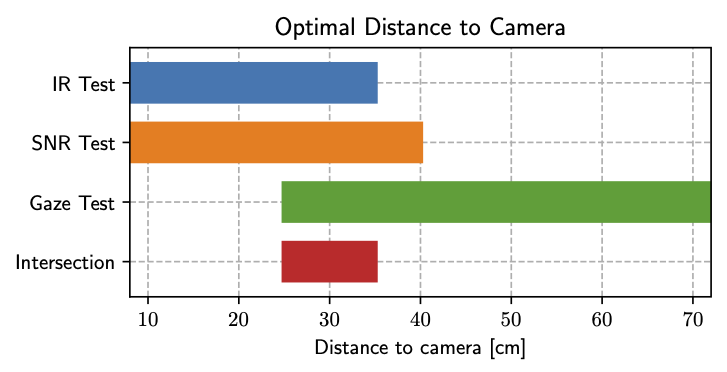}
    \caption{Intervals of optimal distance to the camera according to our experiments.}
    \label{fig:intervals}
\end{figure}

In summary, the above experiments produced three different intervals to optimise different parameters. For good IR performance, the distance to the camera must not exceed 35cm. To get an acceptable SNR, the distance must be below 40cm. To obtain less eye-lead and eyelash obstructions due to the subject's gaze, the distance to the camera has to be above 25cm. Figure~\ref{fig:intervals} illustrates those intervals. The intersection of the three produces the optimal distance to the camera that satisfies the three criteria, and it is the interval between 25 and 35cm. Thus, we calibrated the focal plane of the v1 and v2 camera modules at 30cm. This is an important aspect of the hardware design that is usually overlooked when dealing with IR software.

\subsection{Experiment 5: Eye-Finding Accuracy and Speed}

In this experiment, both segmentation accuracy and the processing speed are evaluated for the first task, which consists of finding eyes in face-ROI images. 
We compared the IoU performance, Eye-tinny-yolo, and two tracking algorithms available in OpenCV \cite{opencv_library}, namely Channel and Spatial Reliability Tracking (CSRT) \cite{lunevzivc2018discriminative} and Kernelized Correlation Filter (KCF) \cite{henriques2012exploiting, danelljan2014adaptive}. It was reported the mean IoU value in the test set $\pm$ is the standard deviation.
It also measured the average inference speed in frames per second (fps) on two computers (PC1 and PC2), a Raspberry-Pi-4B (RasPi) and a Jetson-Nano (Jetson). PC1 has a Ryzen3 prossesor, 16 GB of RAM, and no GPU. PC2 has a Ryzen5 Processor, 16 GB of RAM and a 12 GB GPU. The input image size in this experiment was $800 \times 480$ pixels.

\begin{table*}[!t]
\caption{Evaluation of the Eye Finding Task in terms of IoU Accuracy and Speed.}
\centering
\begin{tabular}{lccccccc}
\hline
\hline
\textbf{Method} & \textbf{N. Param.} & \textbf{\begin{tabular}[c]{@{}c@{}}Resolution \\ {[}px{]}\end{tabular}} & \textbf{IoU} & \textbf{\begin{tabular}[c]{@{}c@{}}PC1 (CPU) \\ {[}fps{]}\end{tabular}} & \textbf{\begin{tabular}[c]{@{}c@{}}PC2 (GPU) \\ {[}fps{]}\end{tabular}} & \textbf{\begin{tabular}[c]{@{}c@{}}RasPi \\ {[}fps{]}\end{tabular}} & \textbf{\begin{tabular}[c]{@{}c@{}}Jetson \\ {[}fps{]}\end{tabular}} \\
\hline
OpenCV CSRT      & NA        & 800$\times$480 & -              & 5.34           & 21.21          & 2.48          & 4.19           \\
OpenCV KCF       & NA        & 800$\times$480 & -              & \textbf{17.14} & 29.14          & \textbf{4.79} & 7.06           \\
Eye-tiny-yolo    & 8,676,244 & 416$\times$416 & \textbf{0.986 ± 0.022} & 3.60   & 42.69          & 1.74          & 4.07           \\
CCNet            & 122,514   & 320$\times$240 & 0.801 ± 0.139  & 6.53           & 22.86          & 1.18          & 8.11           \\
UNet\_xxs (ours) & 28,146    & 160$\times$96  & 0.772 ± 0.135  & 15.66          & \textbf{80.21} & 3.12          & \textbf{16.18} \\
\hline
\hline
\end{tabular}
\label{tab:task_1}
\end{table*}

\begin{table*}[!t]
\caption{Evaluation of the Iris Segmentation Task in terms of IoU Accuracy and Speed}
\centering
\begin{tabular}{lccccccc}
\hline
\hline
\textbf{Method} & \textbf{N. Param.} & \textbf{\begin{tabular}[c]{@{}c@{}}Resolution \\ {[}px{]}\end{tabular}} & \textbf{IoU} & \textbf{\begin{tabular}[c]{@{}c@{}}PC1 (CPU) \\ {[}fps{]}\end{tabular}} & \textbf{\begin{tabular}[c]{@{}c@{}}PC2 (GPU) \\ {[}fps{]}\end{tabular}} & \textbf{\begin{tabular}[c]{@{}c@{}}RasPi \\ {[}fps{]}\end{tabular}} & \textbf{\begin{tabular}[c]{@{}c@{}}Jetson \\ {[}fps{]}\end{tabular}} \\
\hline
DenseNet10       & 210,732 & 320x320 & \textbf{0.946 ± 0.016} & 0.77           & 29.47           & 0.32          & 1.87           \\
CCNet            & 122,514 & 320x240 & 0.919 ± 0.040          & 7.89           & 188.60          & 1.60          & 12.65          \\
UNet\_xxs (ours) & 28,146  & 128x96  & 0.871 ± 0.051          & \textbf{25.78} & \textbf{343.22} & \textbf{3.52} & \textbf{38.47} \\
\hline
\hline
\end{tabular}
\label{tab:task_2}
\end{table*}

\begin{table*}[!t]
\caption{Pupil and Iris Localisation Evaluation}
\centering
\begin{tabular}{llcccc}
\hline
\hline
\textbf{Segmentation Network} & \textbf{Localisation Method} & \textbf{\begin{tabular}[c]{@{}c@{}}Pupil Centre\\ {[}px{]}\end{tabular}} & \textbf{\begin{tabular}[c]{@{}c@{}}Pupil Radius\\ {[}px{]}\end{tabular}} & \textbf{\begin{tabular}[c]{@{}c@{}}Iris Centre\\ {[}px{]}\end{tabular}} & \textbf{\begin{tabular}[c]{@{}c@{}}Iris Radius\\ {[}px{]}\end{tabular}} \\
\hline
DenseNet10       & Centre of Mass      & 0.72 $\pm$ 4.36 & 0.65 ± 0.86 & 1.21 $\pm$ 0.91 & 1.18 $\pm$ 1.30 \\
CCNet            & Mixed (LMS + Hough) & 2.47 $\pm$ 7.83 & 2.20 ± 1.89 & 3.83 $\pm$ 7.53 & 1.96 $\pm$ 1.75 \\
UNet\_xxs (ours) & Mixed (LMS + Hough) & 4.46 $\pm$ 7.71 & 4.50 ± 2.69 & 5.00 $\pm$ 7.49 & 2.89 $\pm$ 2.92 \\
\hline
\hline
\end{tabular}
\label{tab:localisation}
\end{table*}

\begin{table*}[!t]
\caption{Iris Recognition on ND-LG4000-LR}
\centering
\begin{tabular}{cccccc}
\hline
\hline
\textbf{Method} & \textbf{d'} & \textbf{EER} & \textbf{\begin{tabular}[c]{@{}c@{}}FNMR \\ @ FMR=10\%\end{tabular}} & \textbf{\begin{tabular}[c]{@{}c@{}}FNMR \\ @ FMR=1.0\%\end{tabular}} & \textbf{\begin{tabular}[c]{@{}c@{}}FNMR \\ @ FMR=0.1\%\end{tabular}} \\
\hline
DenseNet10        & 5.821 & 0.50\% & 0.22\% & 0.42\% & 0.70\% \\
CCNet             & 4.543 & 1.79\% & 0.79\% & 2.19\% & 4.33\% \\
UNet\_xxs (ours)  & 3.959 & 2.56\% & 0.92\% & 4.35\% & 8.98\% \\
\hline
\hline
\end{tabular}
\label{tab:notredame}
\end{table*}

Table~\ref{tab:task_1} shows the results of this experiment, as well as the input resolutions and the number of parameters used by each method. Because CSRT and KCF are image-processing-based methods, instead of CNN, they do not have trainable parameters. Additionally, the IOU of CSRT and KCF was not computed because they required additional information about the eye's position in the first frame to start tracking. 

In terms of IoU, Eye-tinny-yolo obtained by far the best performance, with 0.986; however, it does so at the expense of using 8 million parameters. CCNet achieved a mean IoU of 0.801 with 112,514 parameters. The mean IoU performance of the proposed UNet\_xxs was 0.772, which is close to that of CCNet but using only 28,146 parameters. A 77.2\% of the accurately recognised area still produces acceptable eye crops, where the entirety of the iris is inside the cropped image. 

In terms of speed, UNet\_xxs is by far the fastest network in all systems and the fastest method for systems with GPU. On the Jetson-Nano, CCNet doubles the speed of Eye-tiny-yolo, and UNet\_xxs doubles that of CCNet. On the Raspberry-Pi, on the other hand, Eye-tiny-yolo is faster than CCNet; however, UNet\_xxs is almost twice as fast as Eye-tiny-yolo.

Finally, the OpenCV algorithm KCF surpassed the speed of UNet\_xxs on systems without a GPU. However, the fact that the captured subject (or biometric attendant) would have to click on the position of their eyes on the first frame makes it impractical on an end-to-end IR system.

\subsection{Experiment 6: Iris Segmentation Accuracy and Speed}

We repeated Experiment 5 for the second task, which is iris segmentation. Speed (in fps) was evaluated using the same systems described in Experiment 5. The IoU metric was computed on the test set using DenseNet10, CCNet and the proposed UNet\_xxs. The image size on the test set was $640 \times 480$ px.

Table~\ref{tab:task_2} shows the input resolution, number of parameters and the results of this experiment. DenseNet10 produced the best IoU of 0.946, using 210,732 parameters. UNet\_xxs obtained an IoU of 0.871, which is high considering that it only used 28,146 parameters to achieve it. Concerning speed, DenseNet10 was far slower than the other two, obtaining speeds below 1 fps in some cases. On the other hand, CCNet and UNet\_xxs were far quicker, with speeds of several fps. On the Jetson-Nano, CCNet obtained 12.65 fps, while UNet\_xxs achieved 38.47. The latter is fast enough for IR frame-by-frame in real-time. Due to their lightweight, CCNet and UNet\_xxs obtained hundreds of fps on a GPU machine. Finally, on the Raspberry-Pi, UNet\_xxs can operate at 3.52 fps, which would produce IR results in a fraction of a second.

\subsection{Experiment 7: Evaluation of Pupil and Iris Localisation}

Once the iris surface has been segmented, the circles that best fit the pupil and iris must be computed. This is essential to obtain the Iris Code \cite{daugman2009iris}. In this experiment, we measured the estimation error when predicting the pupil and iris circles. For the position of the centre of the circles, we measure the error in terms of the Euclidean Distance. On the other hand, we use the Absolute Distance for errors in the estimation of the radius. For predictions of DenseNet10, we used the centre of mass as the localisation algorithm, whereas for CCNet and UNet\_xxs, we used the localisation mixed algorithm of Tapia et al. \cite{tapia2021semantic}. The latter uses LMS when the iris is not occluded by eye leads and Hough when it is occluded \cite{tapia2021semantic}.

Table~\ref{tab:localisation} shows the results of this experiment. DenseNet10's segmentation accuracy undoubtedly helped this method to achieve the best localisation performance with sub-pixel errors. CCNet had the most errors in the 2 px range, except for the iris centre, which was estimated with an average error of 3.83 px. 

Finally, UNet\_xxs produced high estimation errors because the errors are compounded from the segmentation prediction. Those errors are between 4 and 5 px in most cases. However, given that the image size is $640 \times 480$ px, a 5 px error is 1\% of the image height, which, in this context, can be considered a small error.

\subsection{Analysis of Experiments 5, 6 and 7}

Experiments 5, 6 and 7 evaluated the proposed network's speed and accuracy against other state-of-the-art methods. In terms of speed, the proposed UNet\_xxs was the fastest network in both tasks. This is especially true on the Jetson-Nano, which could crop eye images at 16.18 fps and segment them at 38.47 fps. With respect to performance, on the other hand, UNet\_xxs did not perform as well as the other networks. Therefore, for an IR application, we recommend using UNet\_xxs for the first task only and DenseNet10 or CCNet for the second task. In this way, the speed of UNet\_xxs would allow capturing several frames in real-time, cropping the eyes and choosing only the best quality images in terms of sharpness (ISO/IEC 29794-6) for the next process. The second task's speed is not as significant since a single prediction for each eye has to be performed. Therefore, DenseNet10 or CCNet would be better suited to yield results with good accuracy for iris segmentation and localisation.

\subsection{Experiment 8: Iris Recognition Performance}

For a final benchmark evaluation of IR performance, we trained DenseNet10, CCNet and UNet\_xxs on ND-LG4000-LR and performed IR tests using the ICA filters of Czajka et al. \cite{czajka2019domain}. This test can reveal how the segmentation errors propagate and affect IR performance. 

\begin{figure}[!b]
    \centering
    \captionsetup{justification=centering}
    \includegraphics[width=0.75\linewidth]{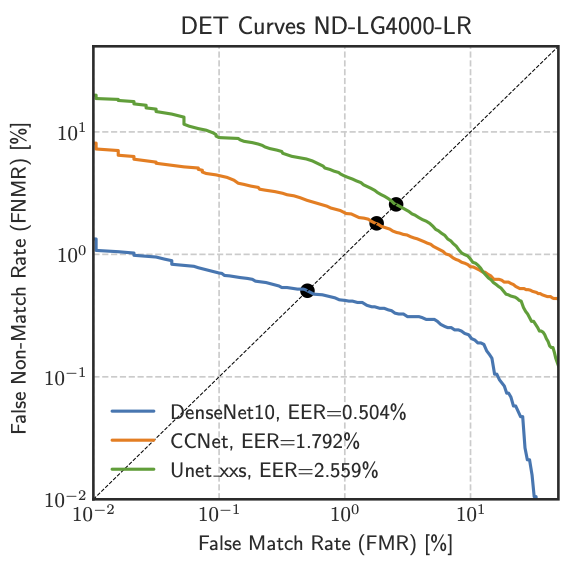}
    \caption{DET curves on ND-LG4000-LR dataset.}
    \label{fig:notredame}
\end{figure}

The DET curves shown in Figure~\ref{fig:notredame} indicate that DenseNet10 has the best overall performance since its DET curve is lower than the others. 
CCNet has better performance than UNet\_xxs for FAR smaller than 15\%. 
However, UNet\_xxs has better performance than CCNet for FAR greater than 15\%.

Table~\ref{tab:notredame} shows the $d'$ and EER of the three networks and the FNMR at three operating points. As in previous tests, DenseNet10 has the best performance. However, the EER of UNet\_xxs is 2.56\%, which is adequately acceptable for real operations. If a 10\% of False Match is permitted in the system, the three networks would have less than a 1\% of False Rejections Rate. On the other hand, when the system allows only a 1\% of FMR, UNet\_xxs  doubles of FNMR of CCNet, at 4.35\%, which is not as desirable. This corroborates that there is a trade-off between speed and performance when using UNet\_xxs.

Finally, note that the EER and FNMR values reported in this work, even for the worst case, are smaller than those reported by Fang et al. \cite{fang2020open} on a different partition on Notre Dame dataset. In that work, CCNet obtained an EER of 4.39\%. This indicates that UNet\_xxs has competitive performance and can be used in a real-world scenario.  

\section{Conclusions}
\label{sec:conclusion}

This work comprehensively analysed the technical aspects of implementing a NIR iris imaging device. The proposed device is lightweight and portable, extracts information from the two eyes, works with NIR illumination, and uses inexpensive single-board computers. We thoughtfully described the factors that affect sensor calibration (iris resolution, SNR and subject's gaze) and how to obtain the best trade-off. The optimal distance to fix the focal plane for the implemented device was 30 $\pm$ 5 cm. 

The necessary processes needed for an end-to-end IR pipeline were also described. The proposed network UNet\_xxs was designed as a lightweight solution for cropping eyes in a face-ROI image and then segmenting the valid iris area in those images. This network achieved the fastest speed among other CNNs for both tasks and reasonable levels of performance. However, we recommend using UNet\_xxs for finding eyes only and DenseNet10 or CCNet for iris segmentation due to their greater accuracy. 

Moreover, for IR we used the ICA BSIF filters implemented on Raspberry-Pi by Fang et al. \cite{fang2020open}. This method normalises and encodes the iris with great performance. For instance, we obtained an EER of 2.56\%, 1.79\% and 0.50\% when segmenting with UNet\_xxs, CCNet and DenseNet10, respectively. Those values are smaller than previous state-of-the-art works.

The described device and methodology make it possible to assemble a NIR imaging device using readily available components. This expands the availability of iris devices at a low cost, which could spark new iris biometrics research in labs with limited resources. Future improvements to the proposed system include adding a PAD stage and using CNN-based iris encoding and IR.

\section{Acknowledgements}

This work is fully supported by the Agencia Nacional de Investigacion y Desarrollo (ANID) through FONDEF IDEA N ID19I10118 and the German Federal Ministry of Education and Research and the Hessen State Ministry for Higher Education, Research and the Arts within their joint support of the National Research Center for Applied Cybersecurity ATHENE.

\bibliography{References}

\bibliographystyle{IEEEtran}

\begin{IEEEbiography}[{\includegraphics[width=1in,height=1.25in,clip,keepaspectratio]{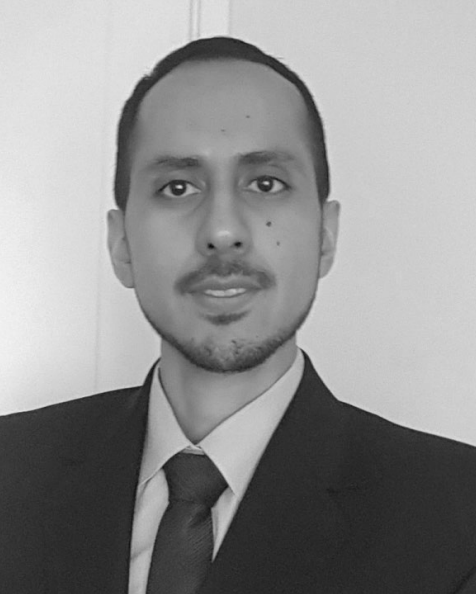}}]
{DANIEL P. BENALCAZAR} (M ‘09) was born in Quito, Ecuador in 1987. He obtained a B.S. in Electronics and Control Engineering from Escuela Politecnica Nacional, Quito, Ecuador, in 2012. He received an M.S. in Electrical Engineering from The University of Queensland, Brisbane, Australia, in 2014, with a minor in Biomedical Engineering. He obtained a PhD in Electrical Engineering from Universidad de Chile, Santiago, Chile in 2020. Ever since, he has participated in various research projects in biomedical engineering and biometrics. From 2015 to 2016, he worked as a Professor at the Central University of Ecuador. He is currently works as a Senior Researcher at the R\&D Center of TOC Biometrics Chile.
\end{IEEEbiography}

\begin{IEEEbiography}
[{\includegraphics[width=1in,height=1.25in,clip,keepaspectratio]{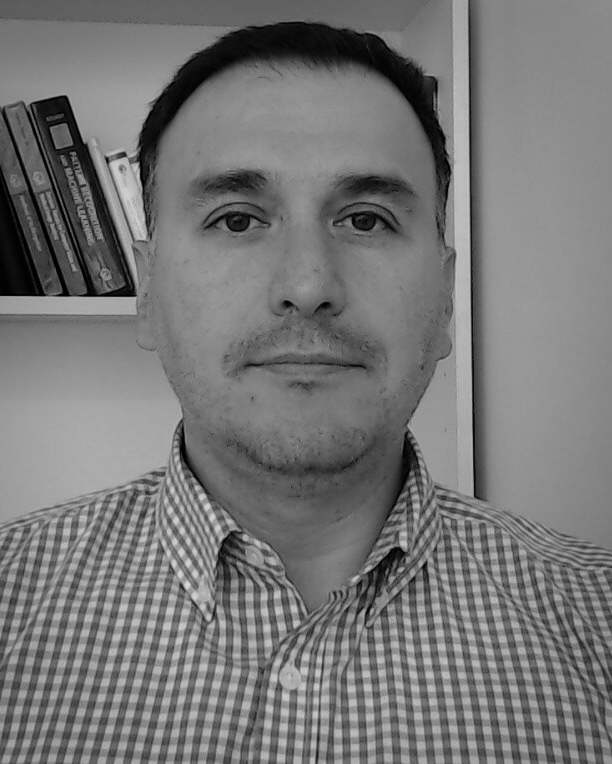}}]{Juan Tapia} received a P.E. degree in Electronics Engineering from Universidad Mayor in 2004, a M.S. in Electrical Engineering from Universidad de Chile in 2012, and a PhD from the Department of Electrical Engineering, Universidad de Chile in 2016. In addition, he spent one year of internship at the University of Notre Dame. In 2016, he received the award for best PhD thesis. From 2016 to 2017, he was an Assistant Professor at Universidad Andres Bello. From 2018 to 2020, he was the R\&D Director for the area of Electricity and Electronics at Universidad Tecnologica de Chile - INACAP. He is currently a Senior Researcher at Hochschule Darmstadt(HDA), and R\&D Director of TOC Biometrics. His main research interests include pattern recognition and deep learning applied to iris biometrics, morphing, feature fusion, and feature selection. 
\end{IEEEbiography}
\vspace{-7pt}

\begin{IEEEbiography}[{\includegraphics[width=1in,height=1.25in,clip,keepaspectratio]{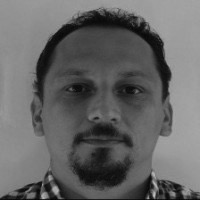}}]{Mauricio Vasquez} received a B.S. in Computer Engineering from Universidad Bio-Bio, Faculty of Informatics, Chile in 2000. His main interests include Computer vision, Graphical interfaces, and UX applied to biometrics.
\end{IEEEbiography}


\begin{IEEEbiography}[{\includegraphics[width=1in,height=1.25in,clip,keepaspectratio]{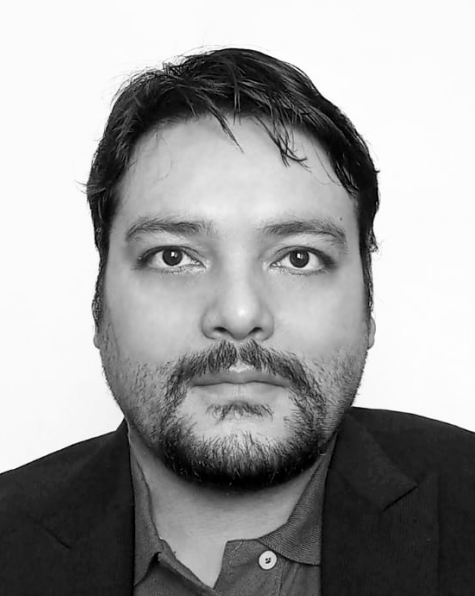}}]{Leonardo Causa} received the P.E. degree in electrical engineering from the Universidad de Chile in 2012, and the M.S. degree in biomedical engineering (BME) from the Universidad de Chile in 2012, he is also finishing the PhD degree in Electrical Engineering and Medical Informatics by cotutelage from U. de Chile and Universite Claude Bernard Lyon 1. His research interests include sleep pattern recognition, signal and image processing, neuro-fuzzy systems applied to the classification of physiological data, and machine and deep learning. He was engaged in research on automated sleep-pattern detection and respiratory signal analysis, fitness for duty, human fatigue, drowsiness, alertness and performance.
\end{IEEEbiography} 
\vspace{-10pt}

\begin{IEEEbiography}[{\includegraphics[width=1.10in,height=1.25in,clip,keepaspectratio]{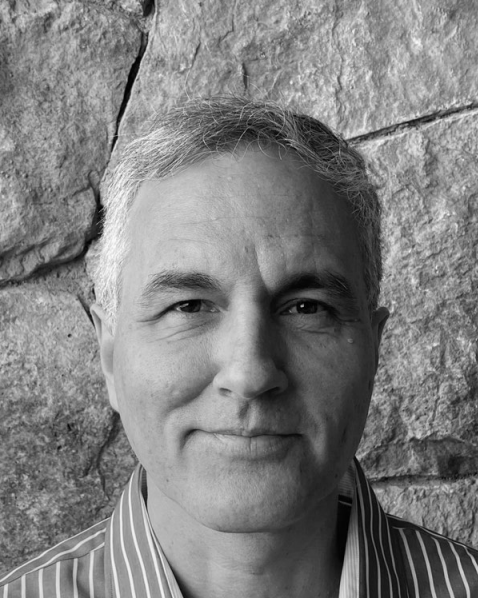}}]{Enrique Lopez Droguett} Droguett is a Professor in the Civil \& Environmental Engineering Department and the Garrick Institute for the Risk Sciences at the University of California, Los Angeles (UCLA), USA, and Associate Editor for both the Journal of Risk and Reliability and the International Journal of Reliability and Safety. He also serves on the Board of Directors of the International Association for Probabilistic Safety Assessment and Management (IAPSAM). Prof. López Droguett conducts research on Bayesian inference and artificial intelligence-supported digital twins and prognostics and health management based on physics-informed deep learning for reliability, risk, and safety assessment of structural and mechanical systems. His most recent focus has been on quantum computing and quantum machine learning for developing solutions for risk and reliability quantification and energy efficiency of complex systems, particularly those involved in renewable energy production. He has led many major studies on these topics for a broad range of industries, including oil and gas, nuclear energy, defence, civil aviation, mining, renewable and hydro energy production and distribution networks. L{ó}pez Droguett has authored more than 250 papers in archival journals and conference proceedings.
\end{IEEEbiography}

\begin{IEEEbiography}[{\includegraphics[width=1in,height=1.25in,clip,keepaspectratio]{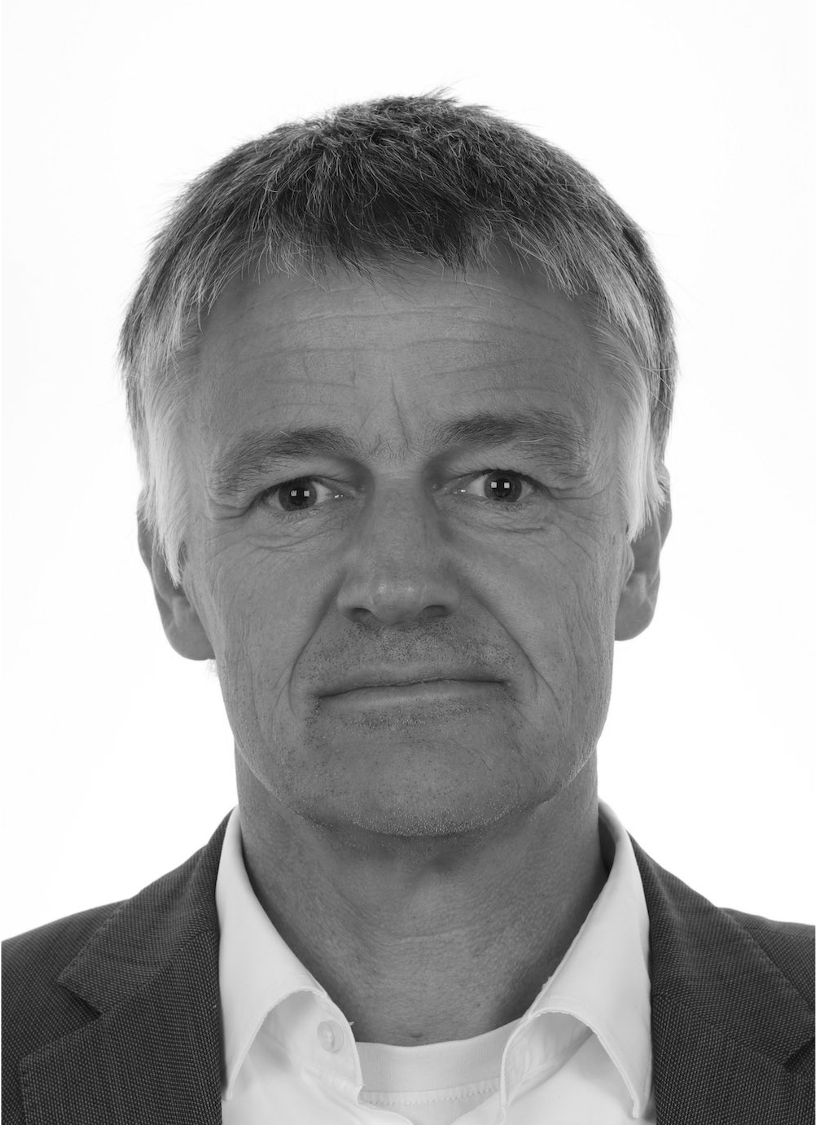}}]{Christoph Busch} is member of the Department of Information Security and Communication Technology (IIK) at the Norwegian University of Science and Technology (NTNU), Norway. He holds a joint appointment with the computer science faculty at Hochschule Darmstadt (HDA), Germany. Further, he has lectured the course Biometric Systems at Denmark’s DTU since 2007. On behalf of the German BSI, he has been the coordinator for the project series BioIS, BioFace, BioFinger, BioKeyS Pilot-DB, KBEinweg and NFIQ2.0. In the European research program, he was the initiator of the Integrated Project 3D-Face, FIDELITY and iMARS. Further, he was/is a partner in the projects TURBINE, BEST Network, ORIGINS, INGRESS, PIDaaS, SOTAMD, RESPECT and TReSPAsS. He is also a principal investigator in the German National Research Center for Applied Cybersecurity (ATHENE). Moreover, Christoph Busch is a co-founder and member of the board of the European Association for Biometrics (www.eab.org) which was established in 2011 and assembled in the meantime more than 200 institutional members. Christoph co-authored more than 500 technical papers and has been a speaker at international conferences. He is a member of the editorial board of the IET journal. 
\end{IEEEbiography}

\end{document}